# Automatic Standardization of Arabic Dialects for Machine Translation

Abidrabbo Alnassan
*Jean Moulin Lyon 3 University*
*Linguistics Research Center - Corpus, Discourse and Societies*
*ILCEA4-CREO (Grenoble Alpes University)*
*abidrabbo.alnassan@univ-lyon3.fr*

## ABSTRACT

*Based on an annotated multimedia corpus, television series Marāyā 2013, we dig into the question of "automatic standardization" of Arabic dialects for machine translation. Here we distinguish between rule-based machine translation and statistical machine translation. Machine translation from Arabic most of the time takes standard or modern Arabic as the source language and produces quite satisfactory translations thanks to the availability of the translation memories necessary for training the models. The case is different for the translation of Arabic dialects. The productions are much less efficient. In our research we try to apply machine translation methods to a dialect/standard (or modern) Arabic pair to automatically produce a standard Arabic text from a dialect input, a process we call "automatic standardization". we opt here for the application of "statistical models" because "automatic standardization" based on rules is more hard with the lack of "diglossic" dictionaries on the one hand and the difficulty of creating linguistic rules for each dialect on the other. Carrying out this research could then lead to combining "automatic standardization" software and automatic translation software so that we take the output of the first software and introduce it as input into the second one to obtain at the end a quality machine translation. This approach may also have educational applications such as the development of applications to help understand different Arabic dialects by transforming dialectal texts into standard Arabic.*

## INTRODUCTION

The Arabic language is a collection of varieties: Standard Arabic (SA) or Classical Arabic (CA), the language used in the Quran as well as in numerous literary texts; Modern Standard Arabic (MSA), the formal and official language of the Arab World; and Arabic dialects (AD), the commonly used informal native varieties. SA differs significantly in its grammatical properties from ADs. ADs have no standard orthographies and rules, they differ from each other and currently have an increasing presence on the web.

Arabic NLP researches, which focused mostly on SA and MSA, is now dealing more with ADs, especially when the DARPA[73] launched, in October 2011, the Broad Operational Language Translation (BOLT) program to attempt to create new techniques for automated translation and linguistic analysis that can be applied to the informal genres of text and speech common in online and in-person communication. Machine translation of ADs is a real challenge because of the lack of dialectal linguistic resources while SA and MSA has a wealth of resources in terms of morphological analyzers, disambiguation systems, annotated data, and parallel corpora.

---

[73] Defense Advanced Research Projects Agency

Based on an annotated multimedia corpus, television series Marāyā 2013, we dig into the question of "automatic standardization" of Arabic dialects for machine translation. Here we distinguish between rule-based machine translation and statistical and neural machine translation. Rule-based machine translation software relies on the use of many linguistic rules and large volumes of dictionary entries for each language pair. The software iterates through the text to be translated and creates an intermediate representation from which the translation is generated. This process requires the use of voluminous dictionaries, syntactic, morphological and semantic data, and numerous linguistic rules.

Statistical machine translation software translates using auto-constructed "statistical models" from monolingual and bilingual corpora. The construction of these "statistical models" requires the prior existence and availability of large volumes of translated texts (translation memories) to train the model to generate the translation.

Neural machine translation, on the other hand, is processed through a neural network where each neuron is a mathematical function that processes data. The initial translation training is done by feeding examples into the neural network and making adjustments based on how much error in the output there was. The network is continually used and continue to fine-tune itself to provide better results.

Machine translation from Arabic most of the time takes standard or modern Arabic as the source language and produces quite satisfactory translations, thanks to the availability of the translation memories necessary for training the models. The case is different for the translation of Arabic dialects. The productions are much less efficient.

In our research we try to apply machine translation methods to a dialect/standard (or modern) Arabic pair to automatically produce a standard Arabic text from a dialect input, a process we call "automatic standardization" versus "automatic translation" or "machine translation". "automatic standardization" is done in one direction: informal or non-standard variety to formal or standard variety of the same language while "automatic translation" is done between different languages regardless of the direction of the process. The other process which goes from the standard variety to the non-standard variety of the same language may be called "automatic destandardization".

We aim through our research to develop a strategy to enrich linguistic resources and parallel ADs-MSA corpora by involving a huge number of human resources not yet involved in this process. We opt here for the application of "statistical models" because "automatic standardization" based on rules is more hard with the lack of standard orthographies for ADs, their numerous varieties, the absence of "diglossic" dictionaries and the difficulty of creating linguistic rules and dedicated tools for each dialect.

## PREVIOUS WORK

Previous research on ADs machine translation has focused on mapping AD input words into MSA equivalents before translating. Researchers have used different techniques to do that. Chiang et al. (2006) built a parser for spoken Levantine Arabic (LA) transcripts using an MSA treebank. They used an LA-MSA lexicon in addition to morphological and syntactic rules to map the LA sentences to MSA. Riesa and Yarowsky (2006) built a statistical morphological segmenter for Iraqi and Levantine speech transcripts, and showed that they outperformed rule-based segmentation with small amounts of training. Abo Bakr et al. (2008) suggested a hybrid, rule-based and statistical, system to map Egyptian Arabic to MSA, using morphological analysis on the input and an Egyptian-MSA lexicon. Sawaf (2010) normalized the dialectal words in a hybrid machine translation system. Salloum and Habash (2011) also mapped AD to MSA.

Some tools exist for preprocessing and tokenizing Arabic text with a focus on ADs. MAGEAD (Habash and Rambow, 2006) is a morphological analyzer and generator that can analyze words into their root/pattern and affixed morphemes, or generate a word in the opposite direction. Amazon's Mechanical Turk (MTurk) help creating annotated resources for computational linguistics. It is an online marketplace that allows "Requesters" to create simple tasks requiring human knowledge, and have them completed by "Workers" from all over the world. Zaidan and Callison-Burch (2011) created the Arabic Online Commentary (AOC) dataset by crawling the websites of three Arabic newspapers[74], and extracting online articles and readers' comments[75]. Over 100k sentences from the AOC were annotated by native Arabic speakers on MTurk to identify ADs and dialect level in each one. The collected labels were used to train automatic dialect identification systems. Laith H. Baniata et al (2018) study the problem of employing a neural machine translation model to translate ADs to MSA. They propose the development of a multitask learning model which shares one decoder among language pairs, and every source language has a separate encoder.

## SPOKEN ARABIC VS STANDARD ARABIC

In spontaneous oral communication, the letter (قَ: qa) of SA is pronounced (أَ: 'a) in some ADs, which can make it difficult, out of context, to determine the word we hear. This difference in pronunciation between spoken Arabic and standard Arabic mobilizes on the one hand the writing of the word produced orally and on the other its meaning in SA.

قَ(qa) → أ('a)
قَلَم(qalam: pen) → أَلَم('alam: pen "in dialect" or pain "in SA")[76]

In some cases, the new word heard is even non-existent in SA.

قَ(qa) → أ('a)
قَرَار(qarār: decision) → أَرَار('arār: decision "in dialect" but there is no meaning to this pronunciation in SA)

A Syrian speaker and a Lebanese or Jordanian speaker can understand each other if both express themselves in their dialectal Arabic, because they are from neighboring linguistic cultures. However, a Syrian and a Moroccan cannot easily communicate through their dialect. There, MSA is essential (Alnassan, 2017). "Automatic standardization" therefore is not only important for the translation of ADs into other foreign languages, but also for creating a passage between the different ADs themselves.

Research on ADs machine translation is mainly based on written productions. The enrichment of machine translation tools therefore also presupposes the use of oral productions.

---

[74] The three newspapers are: 1) Al-Ghad (الغد), a Jordanian newspaper (www.alghad.com), 2) Al-Riyadh (الرياض), a Saudi newspaper (www.alriyadh.com), 3) Al-Youm Al-Sabe' (اليوم السابع), an Egyptian newspaper (www.youm7.com).

[75] The commentary data consists of 3.1M segments, corresponding to 52.1M words.

[76] We follow for the transliteration the system of the Arabica journal : (lettees : ء : ', ب : b, ت : t, ث : t̲, ج : j, ح : ḥ, خ : ḫ, د : d, ذ : ḏ, ر : r, ز : z, س : s, ش : š, ص : ṣ, ض : ḍ, ط : ṭ, ظ : ẓ, ع : ', غ : ġ, ف : f, ق : q, ك : k, ل : l, م : m, ن : n, ه : h, و : w, ي : y. Short vowels : ـَ : a, ـُ : u, ـِ : i. Long vowels : اـ : ā, وـ : ū, يـ : ī).

It is for this reason that our study relied on a television corpus (Marāyā 2013) in which most of the speech is in the Syrian dialect of Damascus.

## MARĀYĀ 2013

Marāyā is a Syrian television series whose first season was broadcast in 1982. Very popular in Syria and the Arab world, this series deals ironically, and sometimes satirically, with themes relating to Syrian daily, social and political life. Marāyā 2013 is the last season of this series. The corpus includes all the thirty episodes broadcast on the Algerian channel Aš-šurūq TV. The average length of each episode is eighteen minutes, while it was around forty minutes in previous seasons. Although the Damascus dialect is the main language used, dialects from other Syrian regions may appear depending on the characters. Similarly, SA and MSA arises from time to time and exceptionally occupies the entirety of episode 25 in which the text is narrative.

### Analysis of the lexicon of Marāyā 2013

The construction and analysis of Marāyā 2013 were initially carried out within the framework of previous research in Arabic language didactics (Alnassan, 2016). The speeches of the characters in the videos were transcribed and annotated using the ELAN (EUDICO Linguistic Annotator)[77] tool. The analysis of the transcripts was implemented in two steps:

- The first consisted of a statistical and descriptive analysis to define the nature of the lexical elements of the corpus and their distribution according to categories (according to lexicon proximity to SA/MSA) and groups (nouns, verbs, adjectives, ..., etc.);
- The second was based on a linguistic analysis (morphological, phonetic and semantic) which aimed to identify useful elements for the improvement of Arabic language teaching manuals.

> The statistical analysis showed that 60% of the lexical elements of the corpus studied are common in SA/MSA and in the Syrian dialect (SD). However, at the level of the construction of the sentence, the phonetic aspect and the semantics in context, the two systems will be even more differentiated.

Through morphological and semantic analysis, we have been able to distinguish:

- Lexical elements that have the same meaning and the same form in SA/MSA and in dialect (أَخِي: ʾḫī: my brother);
- Lexical elements that have undergone slight modifications between SA/MSA and the dialect (letters or/and vowels), retaining the same meaning (كَثِير/كْتِير: kaṯīr/ktīr: many);
- Lexical elements which have the same form in SA/MSA and in dialect but whose meaning is different between the two registers (بُكْرَة: bukrah: "the time just before sunrise" in SA/MSA / "tomorrow" in Syrian dialect).

We also distinguish at the sentence level:

- Constructions which have the same components in SA/MSA and in dialect and which produce the same meaning (إِنْ شَاءَ الله: ʾin šāʾ Allāh: God willing);
- Constructions of lexical elements belonging to SA/MSA, but which are used only in dialectal context (أَمْرِ عِيُونَك: ʾamr ʿyūnak: "as you wish!/ at your service", in the sense of obeying an order or responding to a request kindly.);
- Dialectal constructions containing lexical items related to SA/MSA, retaining the same meaning as the original SA/MSA construction (تِعِدْ لِلْمِيّة أَبِل ما: tʿidd lal-miyyih 'bil mā…:

---

count to a hundred before doing something, the equivalent standard expression is تَعُدُّ لِلمِئَة قَبْلَ أَنْ… : ta'uddu lil-mi'ah qabla 'an…);

- Constructions containing lexical elements related to SA/MSA, but which only exist and have a meaning in dialectal context (تَسْتَوْطِي حِيطِي: tistawṭī ḥiyṭī: "you see it low, my wall/roof"[78], in the sense of challenging someone's contemptuous look at you).

Phonetic analysis allowed us to identify seven letters of SA/MSA that can be pronounced differently in the regional dialect of Damascus. The following table shows, with examples, how to pronounce these letters in different contexts. (Alnassan, 2016b)

**Abbreviations:**
SA: Standard Arabic
L: letter in Arabic script
RdD: Regional dialect of Damascus
T: Transliteration of the Arabic letter
Tra: Translation of the example in English

Table 1. Arabic letters pronounced differently in RdD

| SA letter | | Pronunciation in RdD | | Example | Original word |
|---|---|---|---|---|---|
| L | T | L | T | Word : T : Tra | Word : T |
| ث | t̠ | ث | t̠ | مُثَلَّث : *Mut̠allat̠* : triangle | مُثَلَّث : *Mut̠allat̠* |
| | | ت | t | مِثِل : *mitil* : like | مِثِل : *mit̠il* |
| | | س | s | مَثَلاً : *masalan* : for example | مَثَلاً : *mat̠alan* |
| ذ | d̠ | ذ | d̠ | ذِئْب : *d̠i'b* : wolf | ذِئْب : *d̠i'b* |
| | | ز | z | إِزِن : *'izin* : permission | إِذْن : *'id̠n* |
| | | د | d | إِدِن : *'idin* : hear | أُذُن : *'udun* |
| ص | ṣ | ص | ṣ | صَحِيح : *ṣaḥīḥ* : true | صحيح : *ṣaḥīḥ* |
| | | س | s | صَدِّقْنِي : *saddi'nī* : believe me ! | صَدِّقْنِي : *ṣaddiqnī* |
| ض | ḍ | ض | ḍ | مَرِيض : *marīḍ* : ill | مَرِيض : *marīḍ* |
| | | ز | z | مَزْبُوط : *mazbūṭ* : absolutely | مَضْبُوط : *maḍbūṭ* |
| ظ | ẓ | ظ | ẓ | مُظاهَرَة : *muẓāharah* : demonstration | مُظاهَرَة : *muẓāharah* |
| | | ض | ḍ | ضَهِر : *ḍahr* : back | ظهر : *ẓahr* |
| | | ز | z | غَلِيز : *ġalīz* : heavy / annoying | غَلِيظ : *ġalīẓ* |
| ق | q | ق | q | حَدِيقَة : *ḥadīqah* : garden / park | حَدِيقَة : *ḥadīqah* |
| | | أ | ' | أَدِيم : *'adīm* : ancient | قَدِيم : *qadīm* |
| | | / | g | أَبُو قَاسِم : *'abū gāsim* : the father of Gāsim | أَبُو قَاسِم : *'abū gāsim* |

From these observations, we can imagine the difficulty of carrying out rule-based machine translation or "automatic standardization" of ADs. For this reason, we think that creating and continuously enriching AD-MSA parallel corpora can, with the help of computer tools, considerably advance research in the machine translation of ADs. Audio-visual resources must also be taken into account in this process. Most current research focuses on textual resources.

---

[78] This expression is typically metaphorical. It represents two very distinct situations according to the meanings of the word "حيطي: ḥiyṭī: my wall/my roof" in the Syrian dialect. By the first meaning "حيط: ḥiyṭ: wall", we imagine two neighbors for whom there is a wall separating their house. If one of the two is intrusive and the wall is high, he cannot do anything to disturb his neighbor. On the other hand, if this wall is low, he can easily overcome it and interfere in the affairs of the other. Here, the latter, angry, can use the expression "تَسْتَوْطِي حِيطِي: tistawṭī ḥiyṭī" while they are arguing. In the second case where the meaning is "حيط: ḥiyṭ: roof", the expression refers to a situation where the roof of someone's house is too low so that anyone can climb it and then gain access to the inner courtyard. The metaphor of this expression most often refers to the first image.

## ADs-MSA PARALLEL CORPORA CREATION

Researchers in machine translation of ADs most often use methods that pivot through MSA. Harrath et al. (2017) for example show in a survey eight research works out of thirteen pivoting through MSA to translate ADs into English. Some works are based on open source parallel corpora like what can be found on OPUS[79] (the Open Parallel Corpus). Others build their own corpus.

For our work, we have tried to see if we can find on OPUS an ADs-MSA parallel corpora. By searching on OPUS, we were able to find voluminous resources dealing with the Arabic language. However, almost all the resources found are related to the MSA. The following table shows some of the information obtained by running a query to find Arabic-English parallel corpora. The complete result of our query can be found in the appendix.

Table 2. Some Arabic-English/English-Arabic corpora on OPUS

| Corpus | Arabic-English | | English-Arabic | |
|---|---|---|---|---|
| | Sentence pairs | Words | Sentence pairs | Words |
| United Nations Parallel Corpus | 16,637,291 | 832.98M | 20,044,653 | 904.08M |
| OpenSubtitles v2018 | 25,855,525 | 339.10M | 29,823,188 | 356.14M |
| Tanzil | 184,894 | 13.02M | 187,052 | 13.07M |
| TED2020 v1 | 397,962 | 12.52M | 407,595 | 12.54M |
| tico-19 v2020-10-28 | 3,070 | 0.14M | 3,071 | 0.14M |
| WikiMatrix v1 | 999,763 | 41.98M | 999,763 | 41.98M |
| wikimedia v20210402 | 374,437 | 31.49M | 407,543 | 31.84M |
| Wikipedia | 146,131 | 5.34M | 151,136 | 5.38M |

By searching for corpora containing ADs, we were able to identify two dialects listed among the source languages; the Syrian dialect "ar-SY(Arabic)" and the Tunisian dialect "ar-TN(Arabic)". No dialect has been listed in the target languages. We then made the request to obtain the corpus containing the translations of the Syrian dialect "ar-SY(Arabic)" into MSA (which is represented in OPUS as "ar(Arabic)") and then the corpus containing the translations of the Tunisian dialect "ar-TN(Arabic)" into MSA. The results were as follows:

Table 3. ar-SY(Arabic)-MSA corpora on OPUS

| Search & download resources: | ar_SY (Arabic) | ar (Arabic) | all | ☐ show all versions |
|---|---|---|---|---|

Language resources: click on [ tmx | moses | xces | lang-id ] to download the data! (raw = untokenized, ud = parsed with universal dependencies, alg = word alignments and phrase tables)

| corpus | doc's | sent's | ar tokens | ar_SY tokens | XCES/XML | raw | TMX | Moses | mono | raw | ud | alg | dic | freq | | other files |
|---|---|---|---|---|---|---|---|---|---|---|---|---|---|---|---|---|
| Ubuntu v14.10 | | | | | xces ar ar_SY | ar ar_SY | tmx | moses ar ar_SY | | ar ar_SY | | | dic | ar ar_SY | sample | |
| *total* | 0 | 0 | 0 | 0 | | 0 | | 0 | 0 | | | | | | | |

| color: | | | | | | | | | | | |
|---|---|---|---|---|---|---|---|---|---|---|---|
| size (src+trg): | 16.4k | 32.8k | 65.5k | 0.1M | 0.3M | 0.5M | 1.0M | 2.1M | 4.2M | 8.4M | 16.8M | 33.6M | 67.1M | 134.2M |

Table 4. ar-TN(Arabic)-MSA corpora on OPUS

Search & download resources: | ar_TN (Arabic) | ar (Arabic) | all | ☐ show all versions

Language resources: click on [ tmx | moses | xces | lang-id ] to download the data! (raw = untokenized, ud = parsed with universal dependencies, alg = word alignments and phrase tables)

| corpus | doc's | sent's | ar tokens | ar_TN tokens | XCES/XML | raw | ud | alg | dic | freq | other files |
|--------|-------|--------|-----------|--------------|----------|-----|-----|-----|-----|------|-------------|
| GNOME v1 | 1 | 0.9k | 3.7k | 7.2k | xces ar ar_TN | ar ar_TN | tmx moses ar ar_TN | alg smt | | ar ar_TN | sample |
| total | 1 | 0.9k | 3.7k | 7.2k | | 0.9k | 0.7k 0.9k | | | | |

| color: | | | | | | | | | | | | | | |
|--------|---|---|---|---|---|---|---|---|---|---|---|---|---|---|
| size (src+trg): | 16.4k | 32.8k | 65.5k | 0.1M | 0.3M | 0.5M | 1.0M | 2.1M | 4.2M | 8.4M | 16.8M | 33.6M | 67.1M | 134.2M |

By consulting the tmx version of the two corpora obtained, we were able to discover that the corpus for the Syrian dialect was an empty corpus, while the corpus which was supposed to contain the Tunisian dialect actually contained only words and sentences in MSA accompanied by their equivalent in MSA too. It was therefore not a Tunisian dialectal source language translated into MSA as a target language but rather a source in MSA produced by a Tunisian or in Tunisia and its equivalent in MSA as the target language. Here is an example of the content of this corpus :

Table 5. ar-TN(Arabic)-MSA tmx content

```
64    <tu>
65      <tuv xml:lang="ar"><seg>أنشطة مسلية</seg></tuv>
66      <tuv xml:lang="ar_TN"><seg>اذهب إلى الأنشطة التسلية</seg></tuv>
67    </tu>
68    <tu>
69      <tuv xml:lang="ar"><seg>أنشطة مسلية مختلفة.</seg></tuv>
70      <tuv xml:lang="ar_TN"><seg>أنشطة مسلية و متعددة</seg></tuv>
71    </tu>
72    <tu>
73      <tuv xml:lang="ar"><seg>هندسة</seg></tuv>
74      <tuv xml:lang="ar_TN"><seg>الهندسة</seg></tuv>
75    </tu>
76    <tu>
77      <tuv xml:lang="ar"><seg>الأنشطة الهندسية</seg></tuv>
78      <tuv xml:lang="ar_TN"><seg>الأنشطة الهندسية</seg></tuv>
79    </tu>
80    <tu>
81      <tuv xml:lang="ar"><seg>نشاطات آكل الأعداد</seg></tuv>
82      <tuv xml:lang="ar_TN"><seg>اذهب إلى أنشطة قائمة الأرقام</seg></tuv>
83    </tu>
84    <tu>
85      <tuv xml:lang="ar"><seg>آكل الأعداد هو مجموعة ألعاب حسابية</seg></tuv>
86      <tuv xml:lang="ar_TN"><seg>قائم الأرقام هي ألعاب لممارسة الحساب.</seg></tuv>
87    </tu>
88    <tu>
89      <tuv xml:lang="ar"><seg>استخدام لوحة المفاتيح</seg></tuv>
90      <tuv xml:lang="ar_TN"><seg>استخدام لوحة المفاتيح</seg></tuv>
91    </tu>
```

We also noticed the existence of an "ara (arabic)" in the list of source languages and which does not exist in the list of target languages. Looking also at the tmx version of the parallel corpus for the language pair ara(Arabic)-MSA, we found that these two varieties are only MSA.

Table 6. ara(Arabe)-MSA corpora on OPUS

| Search & download resources: | ara (Arabic) ▾ | | ar (Arabic) ▾ | all ▾ | ☐ show all versions |

Language resources: click on [ tmx | moses | xces | lang-id ] to download the data! (raw = untokenized, ud = parsed with universal dependencies, alg = word alignments and phrase tables)

| corpus | doc's | sent's | ar tokens | ara tokens | XCES/XML | raw | TMX | Moses | mono | raw | ud | alg | dic | freq | other files |
|--------|-------|--------|-----------|------------|----------|-----|-----|-------|------|-----|-----|-----|-----|------|-------------|
| GNOME v1 | 1 | 0.6k | 1.7k | 1.6k | xces ar ara | ar ara | tmx | moses | ar ara | ar ara | | alg smt | | ar ara | sample |
| *total* | 1 | 0.6k | 1.7k | 1.6k | 0.6k | 0.6k | | 0.4k | 0.6k | | | | | | |

| color: | |
|--------|--|
| size (src+trg): | 16.4k 32.8k 65.5k 0.1M 0.3M 0.5M 1.0M 2.1M 4.2M 8.4M 16.8M 33.6M 67.1M 134.2M |

To create their own corpus, some researchers use MTurk (Zaidan and Callison-Burch: 2011a,b; Zbib et al.: 2012). The idea is to create a parallel corpus by hiring non-professional translators and annotators to translate or annotate the sentences that were labeled as being ADs or MSA in documents collected from the web.

This method, from our point of view, is limited because:

- It is based on the work of a small number of contributors (translators or annotators);
- It is costly in terms of financial investment.
- The work is not durable and the enrichment of the corpus is not continuous.
- It is often based on the analysis of written documents.

Our work on Marāyā 2013 pushed us to reflect on methods that may allow us to build big textual corpora based essentially on audio-visual elements. Collecting subtitle texts from films, for example, does not provide this opportunity because in this case the text does not accurately represent the language content of the video. The transcription of the dialogues of Marāyā 2013 series was done manually and took a long time despite the fact that the average duration of each episode was around eighteen minutes. The other seasons of Marāyā had an average duration of forty-five minutes for each episode. We can so imagine how much time and money an individual researcher must spend to manually transcribe the remaining eighteen seasons of Marāyā, where each season contains at least thirty episodes. Automatic video transcription tools (speech to text tools) are not efficient enough for the Arabic language, especially when it comes to ADs.

The creation of textual resources from audio or audio-visual resources cannot therefore be carried out within the framework of individual work, which is our current case. In the same way, creating large ADs-MSA corpora requires the contribution of a very large number of contributors who are able to bring their help to advance this work.

**Development of applications and platforms for the massive transcription and standardization of ADs.**

During our computer-assisted translation (CAT) courses, we invite our students to practice using open source CAT applications and platforms. This practice allows the student to become familiar with these CAT tools. It also contributes to the continuous enrichment of translation memories (TM) which eventually become parallel corpora. The only problem is that these parallel corpora are not accessible to users. In other words, the service is provided free of charge, an individual user can retrieve the TM of his present work, but the parallel corpus produced by all users is not accessible.

To circumvent this problem we imagine the following scenario:

- IT developers or web developers build a web application or a platform allowing the entry of a word, an expression or a sentence in AD, define which AD it belongs to, then standardize it into MSA;
- This web application or platform must be unique and centralized to avoid duplication of data collected by users.

- The application must be accessible for free
- In educational and higher education institutions, we develop introductory and practical courses for the standardization of ADs into MSA.
- The practical part is done using the above-mentioned application or platform.
- Each user can retrieve the result of his current work to be able to develop his own resources;
- Each user can also download the global parallel corpus produced and enriched continuously by all users.
- This possibility of downloading the global parallel corpus may also be available to researchers in DAs and MSA.
- Researchers may also contribute to the development of the application or the platform, or to the development of output evaluation tools, for example.

Through this approach, researchers in dialectology, translation and machine translation of ADs will have an additional resource to those that already exist. A large ADs-MSA parallel corpus, built by a large number of contributors who will not necessarily be translators[80]. If only students from the Arabic language departments of all the universities in the Arab world participate in this work, we will very quickly have the ADs-MSA parallel corpus which we hope to obtain. Such a corpus can help very considerably in the development of the statistical and neural models for ADs machine translation, and before that, the development of statistical and neural models for the automatic standardization of ADs.

In the same way, we can also develop a unique application or platform for the massive transcription of audio and audio-visual resources where the speeches are in ADs. The textual resources obtained by such an approach can be the basis on which the users of the manual standardization application will work.

**CONCLUSION**

Through this brief presentation of the difficulties related to the ADs machine translation, and of the work carried out and in progress in this field, we can come back to the idea that we really need to develop methods and tools to fill the lack of resources for ADs. We have seen that there is currently a significant lack of monolingual dialect corpora based on audio or audio-visual resources. There is also a significant lack in ADs-MSA parallel corpora necessary for training statistical or neural models in ADs machine translation systems.

The solutions we propose: developing a unique application or platform for the massive transcription of audio or audio-visual data which are in ADs, then another application or platform for the standardization of ADs, can significantly help to create and enrich continuously textual resources and large parallel ADs-MSA corpora. We can thus involve a very large number of participants who are not yet involved in this kind of practice while they can help without it being expensive in terms of time and money.

Carrying out this project could then lead to combining "automatic standardization" software and automatic translation software to obtain at the end a quality ADs machine translation.

---

[80] Native speakers of the different ADs, even if they do not know any other foreign language, can participate in this standardization work because SA and MSA are learned in school from childhood.

This approach may also have educational applications such as the development of applications to help understanding different ADs by transforming dialectal texts into standard Arabic.

## Appendix A
### Transliteration of the Arabic alphabet

| Arabic letter | Symbol | Arabic letter | Symbol | Arabic letter | Symbol | Arabic letter | Symbol |
|---|---|---|---|---|---|---|---|
| ء | ʾ | د | d | ض | ḍ | ك | k |
| ب | b | ذ | ḏ | ط | ṭ | ل | l |
| ت | t | ر | r | ظ | ẓ | م | m |
| ث | ṯ | ز | z | ع | ʿ | ن | n |
| ج | j | س | s | غ | ġ | ه | h |
| ح | ḥ | ش | š | ف | f | و | w |
| خ | ḫ | ص | ṣ | ق | q | ي | y |

| Short vowels | | Long vowels | |
|---|---|---|---|
| ـَ | a | ا | ā |
| ـُ | u | و | ū |
| ـِ | i | ي | ī |

## Appendix B
### Arabic-English corpora on OPUS

**Search & download resources:** ar (Arabic) · en (English) · all · ☐ show all versions

**Language resources:** click on [ tmx | moses | xces | lang-id ] to download the data! (raw = untokenized, ud = parsed with universal dependencies, alg = word alignments and phrase tables)

| corpus | doc's | sent's | ar tokens | en tokens | XCES/XML | raw | TMX | Moses | mono | raw | ud | alg | dic | freq | | other files |
|---|---|---|---|---|---|---|---|---|---|---|---|---|---|---|---|
| CCMatrix v1 | 1 | 49.7M | 805.8M | 900.9M | xces ar en | ar en | tmx | moses | ar en | ar en | | | | ar en | sample | |
| WikiMatrix v1 | 1 | 2.0M | 79.6M | 1.0G | xces ar en | ar en | tmx | moses | ar en | ar en | | | | ar en | sample | |
| UNPC v1.0 | 114067 | 16.6M | 394.7M | 445.4M | xces ar en | ar en | tmx | moses | ar en | ar en | | alg | | ar en | sample | |
| MultiUN v1 | 67617 | 8.2M | 201.7M | 228.2M | xces ar en | ar en | tmx | moses | ar en | ar en | | alg | | ar en | query sample | |
| CCAligned v1 | 507 | 13.0M | 188.7M | 200.6M | xces ar en | ar en | tmx | moses | ar en | ar en | | | | ar en | sample | |
| wikimedia v20210402 | 1 | 0.4M | 24.7M | 349.2M | xces ar en | ar en | tmx | moses | ar en | ar en | | | | ar en | sample | |
| OpenSubtitles v2018 | 8256 | 4.6M | 26.8M | 29.9M | xces ar en | ar en | tmx | moses | ar en | ar en | | alg smt | dic | ar en | sample | xces/alt |
| XLEnt v1.1 | 1 | 5.6M | 19.1M | 18.7M | xces ar en | ar en | tmx | moses | ar en | ar en | | | | ar en | sample | |
| QED v2.0a | 5033 | 0.7M | 6.6M | 9.5M | xces ar en | ar en | tmx | moses | ar en | ar en | | alg smt | dic | ar en | sample | |
| TED2020 v1 | 3879 | 0.4M | 6.4M | 8.1M | xces ar en | ar en | tmx | moses | ar en | ar en | | | | ar en | sample | |
| Tanzil v1 | 30 | 0.2M | 7.9M | 5.6M | xces ar en | ar en | tmx | moses | ar en | ar en | | alg smt | | ar en | sample | |
| News-Commentary v16 | 7185 | 83.2k | 5.0M | 3.8M | xces ar en | ar en | tmx | moses | ar en | ar en | | alg smt | | ar en | sample | |
| UN v20090831 | 1 | 74.1k | 3.3M | 3.7M | xces ar en | ar en | tmx | moses | ar en | ar en | | | | ar en | query sample | |
| Wikipedia v1.0 | 1 | 0.2M | 3.2M | 3.5M | xces ar en | ar en | tmx | moses | ar en | ar en | | alg smt | dic | ar en | query sample | |
| TED2013 v1.1 | 1 | 0.2M | 2.4M | 3.0M | xces ar en | ar en | tmx | moses | ar en | ar en | | alg smt | | ar en | sample | |
| GNOME v1 | 1313 | 0.5M | 2.4M | 2.6M | xces ar en | ar en | tmx | moses | ar en | ar en | | | | ar en | sample | |
| bible-uedin v1 | 2 | 61.5k | 0.9M | 1.5M | xces ar en | ar en | tmx | moses | ar en | ar en | | alg smt | dic | ar en | sample | |
| GlobalVoices v2018q4 | 3875 | 58.3k | 1.0M | 1.3M | xces ar en | ar en | tmx | moses | ar en | ar en | | alg smt | dic | ar en | sample | |
| KDE4 v2 | 784 | 0.1M | 0.7M | 0.8M | xces ar en | ar en | tmx | moses | ar en | ar en | | alg smt | dic | ar en | sample | |
| Mozilla-I10n v1 | 1 | 51.7k | 0.2M | 0.7M | xces ar en | ar en | tmx | moses | ar en | ar en | | | | ar en | sample | |
| ELRC_2922 v1 | 1 | 15.1k | 0.3M | 0.2M | xces ar en | ar en | tmx | moses | ar en | ar en | | | | ar en | sample | |
| EUbookshop v2 | 30 | 1.7k | 80.0k | 0.4M | xces ar en | ar en | tmx | moses | ar en | ar en | | alg smt | | ar en | query sample | moses/strict |
| infopankki v1 | 290 | 16.0k | 0.2M | 0.2M | xces ar en | ar en | tmx | moses | ar en | ar en | | alg smt | dic | ar en | sample | |
| Tatoeba v2022-03-03 | 1 | 27.3k | 0.1M | 0.2M | xces ar en | ar en | tmx | moses | ar en | ar en | | | | ar en | sample | |
| tico-19 v2020-10-28 | 1 | 3.1k | 67.9k | 70.4k | xces ar en | ar en | tmx | moses | ar en | ar en | | alg smt | dic | ar en | sample | |
| Ubuntu v14.10 | 1 | | | | xces ar en | ar en | tmx | moses | ar en | ar en | | | dic | ar en | sample | |
| *total* | 212879 | 102.8M | 1.8G | 3.3G | | | 102.8M | | 101.0M | 115.6M | | | | | | |

**color:**

**size (src+trg):** 16.4k 32.8k 65.5k 0.1M 0.3M 0.5M 1.0M 2.1M 4.2M 8.4M 16.8M 33.6M 67.1M 134.2M

## English-Arabic corpora on OPUS

Search & download resources: [ en (English) ▾ ]  [ ar (Arabic) ▾ ]  [ all ▾ ]  ☐ show all versions

Language resources: click on [ tmx | moses | xces | lang-id ] to download the data! (raw = untokenized, ud = parsed with universal dependencies, alg = word alignments and phrase tables)

| corpus | doc's | sent's | ar tokens | en tokens | XCES/XML | raw | TMX | Moses | mono | raw | ud | alg | dic | freq | | other files |
|---|---|---|---|---|---|---|---|---|---|---|---|---|---|---|---|---|
| CCMatrix v1 | 1 | 49.7M | 805.8M | 900.9M | xces ar en | ar en | tmx | moses | ar en | ar en | | | | ar en | sample | |
| WikiMatrix v1 | 1 | 2.0M | 79.6M | 1.0G | xces ar en | ar en | tmx | moses | ar en | ar en | | | | ar en | sample | |
| UNPC v1.0 | 114067 | 16.6M | 394.7M | 445.4M | xces ar en | ar en | tmx | moses | ar en | ar en | | alg | | ar en | sample | |
| MultiUN v1 | 67617 | 8.2M | 201.7M | 228.2M | xces ar en | ar en | tmx | moses | ar en | ar en | | alg | | ar en | query sample | |
| CCAligned v1 | 507 | 13.0M | 188.7M | 200.6M | xces ar en | ar en | tmx | moses | ar en | ar en | | | | ar en | sample | |
| wikimedia v20210402 | 1 | 0.4M | 24.7M | 349.2M | xces ar en | ar en | tmx | moses | ar en | ar en | | | | ar en | sample | |
| OpenSubtitles v2018 | 8256 | 4.6M | 26.8M | 29.9M | xces ar en | ar en | tmx | moses | ar en | ar en | | alg smt | dic | ar en | query sample | xces/alt |
| XLEnt v1.1 | 1 | 5.6M | 19.1M | 18.7M | xces ar en | ar en | tmx | moses | ar en | ar en | | | | ar en | sample | |
| QED v2.0a | 5033 | 0.7M | 6.6M | 9.5M | xces ar en | ar en | tmx | moses | ar en | ar en | | alg smt | dic | ar en | sample | |
| TED2020 v1 | 3879 | 0.4M | 6.4M | 8.1M | xces ar en | ar en | tmx | moses | ar en | ar en | | | | ar en | query sample | |
| Tanzil v1 | 30 | 0.2M | 7.9M | 5.6M | xces ar en | ar en | tmx | moses | ar en | ar en | | alg smt | dic | ar en | sample | |
| News-Commentary v16 | 7185 | 83.2k | 5.0M | 3.8M | xces ar en | ar en | tmx | moses | ar en | ar en | | alg smt | dic | ar en | sample | |
| UN v20090831 | 1 | 74.1k | 3.3M | 3.7M | xces ar en | ar en | tmx | moses | ar en | ar en | | alg smt | dic | ar en | sample | |
| Wikipedia v1.0 | 1 | 0.2M | 3.2M | 3.5M | xces ar en | ar en | tmx | moses | ar en | ar en | | alg smt | dic | ar en | query sample | |
| TED2013 v1.1 | 1 | 0.2M | 2.4M | 3.0M | xces ar en | ar en | tmx | moses | ar en | ar en | | alg smt | dic | ar en | query sample | |
| GNOME v1 | 1313 | 0.5M | 2.4M | 2.6M | xces ar en | ar en | tmx | moses | ar en | ar en | | alg smt | dic | ar en | sample | |
| bible-uedin v1 | 2 | 61.5k | 0.9M | 1.5M | xces ar en | ar en | tmx | moses | ar en | ar en | | alg smt | dic | ar en | sample | |
| GlobalVoices v2018q4 | 3875 | 58.3k | 1.0M | 1.3M | xces ar en | ar en | tmx | moses | ar en | ar en | | alg smt | dic | ar en | query sample | |
| KDE4 v2 | 784 | 0.1M | 0.7M | 0.8M | xces ar en | ar en | tmx | moses | ar en | ar en | | alg smt | dic | ar en | query sample | |
| Mozilla-I10n v1 | 1 | 51.7k | 0.2M | 0.7M | xces ar en | ar en | tmx | moses | ar en | ar en | | alg smt | dic | ar en | sample | |
| ELRC_2922 v1 | 1 | 15.1k | 0.3M | 0.3M | xces ar en | ar en | tmx | moses | ar en | ar en | | alg smt | dic | ar en | sample | |
| EUbookshop v2 | 30 | 1.7k | 80.0k | 0.4M | xces ar en | ar en | tmx | moses | ar en | ar en | | alg smt | dic | ar en | query sample | moses/strict |
| infopankki v1 | 290 | 16.0k | 0.2M | 0.2M | xces ar en | ar en | tmx | moses | ar en | ar en | | alg smt | dic | ar en | sample | |
| Tatoeba v2022-03-03 | 1 | 27.3k | 0.1M | 0.2M | xces ar en | ar en | tmx | moses | ar en | ar en | | | | ar en | sample | |
| tico-19 v2020-10-28 | 1 | 3.1k | 67.9k | 70.4k | xces ar en | ar en | tmx | moses | ar en | ar en | | | dic | ar en | sample | |
| Ubuntu v14.10 | 1 | 0.2M | | | xces ar en | ar en | tmx | moses | ar en | ar en | | | dic | ar en | sample | |
| *total* | 212879 | 102.8M | 1.8G | 3.3G | | 102.8M | | 101.0M | 115.6M | | | | | | | |

| color: | | | | | | | | | | | | |
|---|---|---|---|---|---|---|---|---|---|---|---|---|
| size (src+trg): | 16.4k | 32.8k | 65.5k | 0.1M | 0.3M | 0.5M | 1.0M | 2.1M | 4.2M | 8.4M | 16.8M | 33.6M | 67.1M | 134.2M |